\documentclass[10pt,twocolumn,letterpaper]{article}

\usepackage{times}
\usepackage{epsfig}
\usepackage{graphicx}
\usepackage{amsmath}
\usepackage{amssymb}
\usepackage{multirow}
\usepackage{tabularx}
\usepackage{authblk}

\usepackage{color,xcolor}
\usepackage{epsfig}
\usepackage{graphicx}

\usepackage{adjustbox}
\usepackage{array}
\usepackage{booktabs}
\usepackage{colortbl}
\usepackage{float,wrapfig}
\usepackage{hhline}
\usepackage{multirow}
\usepackage{subcaption} 

\usepackage{amsmath,amsfonts,amsthm,amssymb}
\usepackage{bm}
\usepackage{nicefrac}
\usepackage{microtype}

\usepackage{changepage}
\usepackage{extramarks}
\usepackage{fancyhdr}
\usepackage{lastpage}
\usepackage{setspace}
\usepackage{soul}
\usepackage{xspace}

\usepackage{url}

\usepackage{algorithm, algorithmic}
\usepackage{enumerate}

\newcolumntype{L}[1]{>{\raggedright\let\newline\\\arraybackslash\hspace{0pt}}m{#1}}
\newcolumntype{C}[1]{>{\centering\let\newline\\\arraybackslash\hspace{0pt}}m{#1}}
\newcolumntype{R}[1]{>{\raggedleft\let\newline\\\arraybackslash\hspace{0pt}}m{#1}}

\newcommand{\sect}[1]{Section~\ref{#1}}

\newcommand{\fig}[1]{Figure~\ref{#1}}
\newcommand{\tbl}[1]{Table~\ref{#1}}

\newcommand{\ignore}[1]{}

\renewcommand*{\thefootnote}{\fnsymbol{footnote}}

\makeatletter
\DeclareRobustCommand\onedot{\futurelet\@let@token\@onedot}
\def\@onedot{\ifx\@let@token.\else.\null\fi\xspace}

\def\eg{\emph{e.g}\onedot} 
\def\ie{\emph{i.e}\onedot} 
 
\def\etc{\emph{etc}\onedot} 
 
\def\etal{\emph{et al}\onedot}
\makeatother

\definecolor{MyDarkBlue}{rgb}{0,0.08,1}
\definecolor{MyDarkGreen}{rgb}{0.02,0.6,0.02}
\definecolor{MyDarkRed}{rgb}{0.8,0.02,0.02}
\definecolor{MyDarkOrange}{rgb}{0.40,0.2,0.02}
\definecolor{MyPurple}{RGB}{111,0,255}
\definecolor{MyRed}{rgb}{1.0,0.0,0.0}
\definecolor{MyGold}{rgb}{0.75,0.6,0.12}
\definecolor{MyDarkgray}{rgb}{0.66, 0.66, 0.66}

\newcommand{\mysubsubsection}[1]{\vspace{0.1cm} \noindent {\bf #1}:}

\def\@maketitle
{
  \newpage
  \null
  \vskip .375in
  \begin{center}
    {\Large \bf \@title \par}
    \vspace*{24pt}
    {
      \large
      \lineskip .5em
      \begin{tabular}[t]{c}
        \@author\\
      \end{tabular}
      \par
    }
    \vskip .5em
    \vspace*{12pt}
  \end{center}
}

\def\abstract
{%
  \centerline{\large\bf Abstract}%
  \vspace*{12pt}%
  \it%
}

\usepackage[pagebackref=true,breaklinks=true,colorlinks,bookmarks=false]{hyperref}

\begin{document}

\title{\vspace{-5ex}Deep Multi-Modal Image Correspondence Learning}


\author[1]{Chen Liu}
\author[2]{Jiajun Wu}
\author[3]{Pushmeet Kohli}
\author[1]{Yasutaka Furukawa}
\affil[1]{Washington University in St. Louis}
\affil[2]{Massachusetts Institute of Technology}
\affil[3]{Microsoft Research}
\renewcommand\Authsep{\quad\quad}
\renewcommand\Authand{\quad\quad}
\renewcommand\Authands{\quad\quad}
\date{\vspace{-0ex}}
\renewcommand{\thefootnote}{\arabic{footnote}}

\twocolumn[{%
  \renewcommand\twocolumn[1][]{#1}%
  \maketitle
  \begin{center}
    \vspace{-10pt}
    \includegraphics[width=0.9\textwidth]{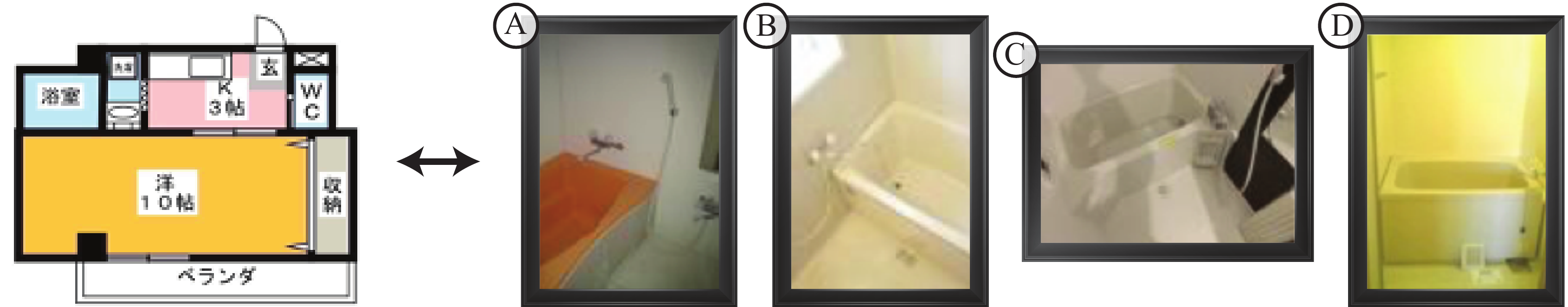}
    \captionof{figure}{Which of the four photographs corresponds to the
      left floorplan?
      The task requires careful and sophisticated human reasoning, unlike
      many other computer vision problems that only require instant human
      attention.
      This paper explores the potential of deep-neural networks in solving
      such a problem.
      The answer is in the footnote\color{red}$^2$\color{black}.}
    \vspace{10pt}
    \label{fig:teaser}
  \end{center}%
}]

\maketitle

\begin{abstract}
  
  
Inference of correspondences between images from different modalities is
an extremely important perceptual ability that enables humans to
understand and recognize cross-modal concepts. In this paper, we
consider an instance of this problem that involves matching photographs
of building interiors with their corresponding floorplan. This is a
particularly challenging problem because a floorplan, as a stylized
architectural drawing, is very different in appearance from a
color photograph. Furthermore, individual photographs by themselves
depict only a part of a floorplan (\eg, kitchen, bathroom, and living
room). We propose the use of a number of different neural network architectures
for this task, which are trained and evaluated on a novel large-scale
dataset of 5 million floorplan images and 80 million associated
photographs. Experimental evaluation reveals that our neural network
architectures are able to identify visual cues that result in reliable
matches across these two quite different modalities. In fact, the
trained networks are able to even outperform human subjects in several
challenging image matching problems. Our result implies that neural
networks are effective at perceptual tasks that require long periods of
reasoning even for humans to solve\footnote[1]{Project page: http://www.cse.wustl.edu/${\sim}$chenliu/floorplan-matching.html}.

\end{abstract}

\section{Introduction}

Our world is full of imagery rich in modalities. Product manuals utilize
stylized line drawings to emphasize product features.  Building
blueprints are precise technical drawings for construction design and
planning.
\footnotetext[2]{\rotatebox[origin=c]{180}{A is the answer.}}
Artists use characteristics brush-strokes for unique aesthetic
appeals.
Humans learn to understand images of different modalities and associate them
with the physical views through our eyes.
%
%

In a quest to reach the level of human visual intelligence, an important
capability for computer vision is to associate images in different
modalities.
Surprisingly, apart from some notable exceptions~\cite{wang2015lost,castrejonlearning}, cross-modal image association has been an under-explored
problem in the field, while cross-modal data analysis has recently been
popular, for instance, between an image and natural language~\cite{karpathy2015deep} or audio~\cite{aytar2016soundnet,owens2016ambient}. 

In this paper, we learn to match a floorplan image and photographs of
building interiors.  We use a new large-scale database of 5 million
floorplan images of residential units (mostly apartments) associated
with 80 million photographs~\cite{homes}.
This is a challenging cross-modal image correspondence learning problem
because 1) a floorplan is a stylized architectural drawing, only
capturing rough geometric structure in an orthographic view, which is
very different from a photograph; and 2) only a part of a floorplan
image corresponds to each photograph (\eg, kitchen, bathroom, and living
room).

We formulate and solve several variants of the cross-modal
matching problem (see \fig{fig:teaser}). Specifically, we first consider the case
where a single photograph (of known or unknown room type -
living room, bathroom, bedroom \etc)  needs to be matched to
the floorplan.  Thereafter, we consider the case where a set
of photographs corresponding to different parts of the
building need to be matched to the floorplan. The latter
variant requires us to explore the use of different neural
architectures for handling unordered sets of
data-points. Our experimental results show that our models
are extremely effective in discovering visual cues that
associate floorplan images and photographs across vastly
different modalities. In fact, our models outperform human
subjects, who often need half a minute or more to solve each
matching test.

The key contributions of this paper are: 1) we introduce a set of
challenging multi-modal image-set matching problems to the community
with human baselines obtained through Amazon Mechanical Turk (AMT); 2)
we develop deep neural network based methods that outperform humans in
all these tasks; 3) we analyze the behavior of different models by
generating visualizations that provide intuitions as to how they reason
the problem; and 4) we present applications enabled by this new
capability to perform cross-modal matching between floorplans and
photographs.

\section{Related work}
\label{section:related}



\mysubsubsection{Image correspondence} Image correspondence has a long
history in vision. While feature detection and matching techniques based
on descriptors such as SIFT~\cite{lowe1999object} and
SURF~\cite{bay2006surf} have been successful for narrow-baseline
problems, they perform poorly for large-baseline problems. To overcome
these problems, researchers have proposed detecting and matching
larger-scale structure such as building facades for ground-to-aerial
image matching~\cite{bansal2012ultra,wolff2016regularity}, or to learn
the relationships of features across large-baselines (\eg, ground to
satellite images)~\cite{lin2013cross}.

%
%
%

%

More recently, a number of neural network architectures have also been proposed for the problem.
Long \etal used CNN to improve feature localization~\cite{long2014convnets}.
Siamese networks~\cite{bromley1993signature} and their variants have also become a popular
architecture to learn distance
metrics~\cite{han2015matchnet,zagoruyko2015learning,lin2015learning}
between local image patches.
Altwaijry \etal~\cite{altwaijry2016learning} extended the Siamese
architecture with spatial transformer
network~\cite{jaderberg2015spatial} to perform feature detection as well
as matching inside a network. All these methods focus on comparing
images in a single modality (\ie, photographs). Our problem is
fundamentally different in that we need to handle images in vastly
different styles (floorplans vs. photographs) and viewpoints, and to
reason with sets of different cardinalities.

\mysubsubsection{Multi-modal matching}
Multi-modal data analysis is becoming an increasingly active area of
research. In the field of computer vision, Visual Question Answering is a notable
example~\cite{antol2015vqa}, where a model needs to jointly reason about
the image and a natural language question to produce an answer.  Image captioning has been another popular
problem involving images and natural language~\cite{bernardi2016automatic}. Visual storytelling is its
natural extension, where the task is to write a story given a set of
images~\cite{huang2016visual}. All these studies focus on multi-modal
data analysis between images and a language, while we focus on
multi-modal image analysis.

The closest work to ours is the cross-modal representation learning by
Castrej\'{o}n \etal~\cite{castrejonlearning}. They learn common scene
representation across five different modalities, in particular,
photographs, clip-arts, sketches, texts, and ``spatial-text''. The key
difference from our work is that they focus on classifying scene
categories. Instead, we seek to classify  instances of indoor
scenes, requiring more precise geometric reasoning and content
analysis. Furthermore, samples in their work are in one-to-one
correspondence across modalities, and are spatially aligned.  In our
work, multiple samples in one modality as a whole (photographs)
correspond to only one sample in the other (floorplan). This prohibits
standard cross-modal representation learning as a single photograph is
not equivalent to a single floorplan.

\mysubsubsection{Indoor scene understanding}
Matching photographs and floorplans has been an important problem in
the context of indoor scene localization. Most existing techniques
employ explicit geometry reasoning. For instance, Chu
\etal~\cite{chu2015you} align visual odometry reconstruction against
a floorplan.  In addition to multi-view geometry, Martin
\etal~\cite{martin20143d} exploit the order of photographs taken in
an individual photo album to align photographs against a floorplan.
Wang \etal~\cite{wang2015lost} enable the alignment of a single image
against a floorplan, via more sophisticated image understanding
techniques involving scene layout, store boundaries, and texts.
Liu \etal~\cite{liu2015rent3d} employ similar image processing
techniques to align a photograph to a floorplan of a residential unit.
%
In all these works, a ground-truth floorplan is given, and the problem
is to perform image localization via hand-coded features. In contrast, our
work studies machine vision's capability to automatically learn an
effective representation that allows us to compare images of quite
different modalities, floorplans and photographs.





\section{Cross-modal image matching problem}
\label{section:problem}
\begin{figure}[t]
  \includegraphics[width=\columnwidth]{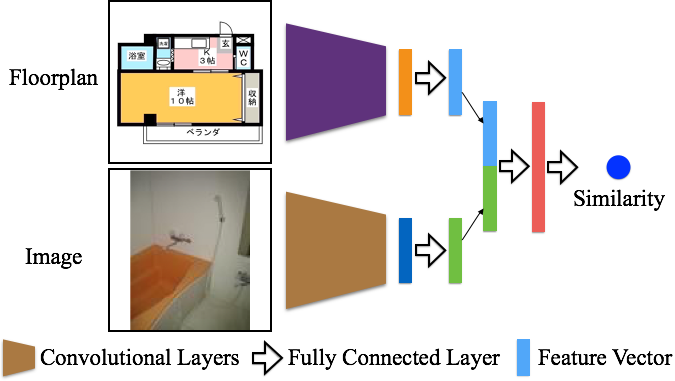}
  \caption{We use a Siamese network architecture for pair matching, with its two arms for learning representations for floorplans and photographs, respectively. The network then predicts the confidence of whether a pair belong to the same apartment. }
  \label{figure:binary}
\end{figure}

\begin{figure*}[t]
  \includegraphics[width=\linewidth]{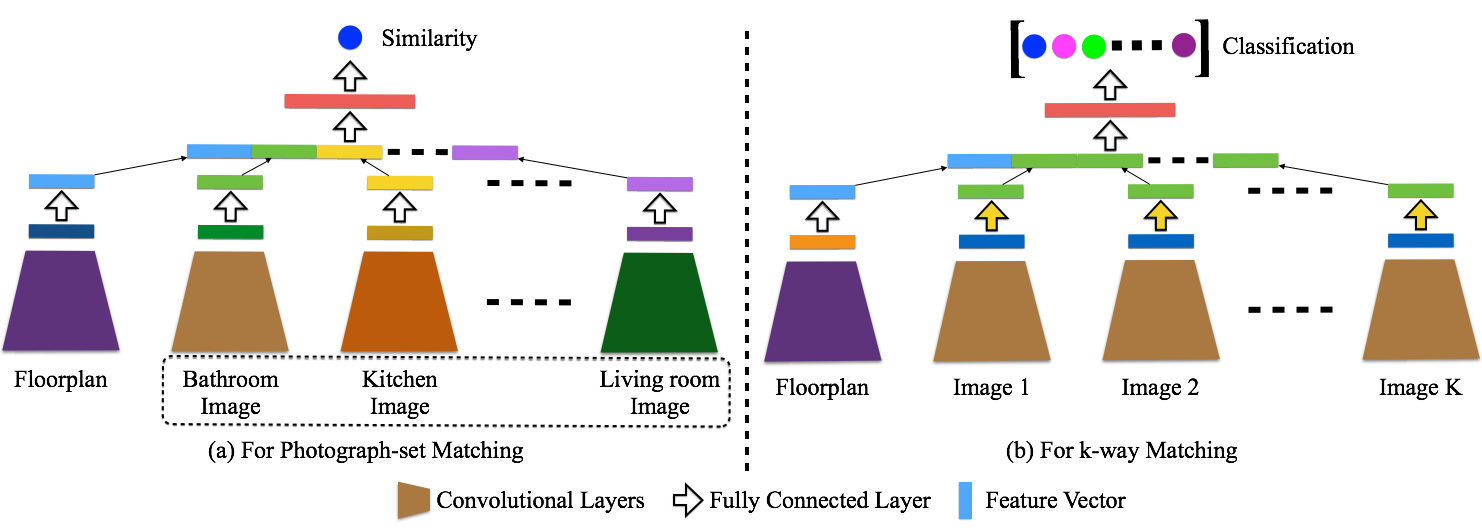}
  \caption{{\bf Left}: our
    network architecture for matching one floorplan against multiple
    monocular images with different room types from the same apartment (photograph-set matching). 
    {\bf Right}: our
    network architecture for matching one floorplan against multiple
    monocular images from different apartments ($k$-way matching).}
  \label{figure:room}
\end{figure*}

This paper explores a diverse set of matching
problems between floorplans and photographs, as shown in Figures \ref{figure:binary} and \ref{figure:room}.  
The basic problem configuration is to provide a model with one floorplan
image and one photograph, and ask whether they come from the same
apartment. For a comprehensive study, we investigate more problem
variations by considering the following three problem settings.


\vspace{0.2cm}
\noindent $\bullet$ First, we vary the number of photographs for each
apartment. For example, we may supply a bathroom photograph only, or
bathroom, kitchen, and living-room photographs altogether.

\vspace{0.2cm}
\noindent $\bullet$ 
Second, we vary the number $(k)$ of apartments or matching candidates. When $k=1$, the model essentially answers a ``Yes / No'' question --- if the floorplan matches the photograph or not. When $k\ge2$, the model must choose the photograph that matches the floorplan from multiple choices.

\vspace{0.2cm}
\noindent $\bullet$ Third, we explore both room-aware and room-agnostic
matching. Suppose we are to match a floorplan against a set of three
photographs. In a case of room-aware matching, the network knows the
room type for each photograph, and can train room-type specific network
modules. In a case of room-agnostic matching, the network is given
randomly ordered photographs without their room type information.


\section{Neural cross-modal image matching}
\label{section:models}

This section proposes our neural approach to the diverse family of
cross-modal image matching problems. We
provide details for some representative problem configurations; it
is straightforward to construct the architecture for the remaining
ones.

\subsection{Pair matching}


This is the basic configuration. Given one floorplan image and one
photograph, a network predicts if these two come from the same
apartment. We formulate this as a similarity regression problem, where
the output score ranges from -1 to 1.
Inspired by the recent success of correspondence matching
approaches~\cite{zagoruyko2015learning, han2015matchnet,
simo2015discriminative, lin2015learning, altwaijry2016learning},
we form a Siamese network followed by a fully connected regression
network. The two arms of the Siamese network learn a feature 
representation of floorplans and photographs, respectively.
We show the network structure in \fig{figure:binary}.
In a room-aware setting, we train a room-type specific encoder inside
the Siamese network. In a room-agnostic setting, we train a single
photograph encoder regardless of the room types.

We initialize each encoder with VGG16~\cite{simonyan2014very}, while
changing the output feature dimension of ${\it fc6}$ to 512 and
removing $ {\it fc7}$ and ${\it fc8}$.
%
%
%
%
The regression network consists of two fully connected
layers. The first takes the concatenation of two feature vectors from
the Siamese arms and outputs a 1,024 dimensional vector. The second
layer regresses the similarity score.
We follow~\cite{zagoruyko2015learning} and use a hinge loss.

\subsection{Photograph-set matching}
\label{section:set-matching-models}

As multiple photographs provide more cues in improving the matching accuracy,
%
we consider the matching problem between a floorplan and a set of
photographs of an apartment.

%
Suppose we have a set of $n$ photographs (\eg, bathroom, kitchen, or
living-room). A feature vector from the floorplan encoder and $n$
feature vectors from the photograph encoders are concatenated into fully
connected layers. \fig{figure:room}a shows our architecture for this
problem.

%
%
In a room-aware setting, we always pass a set of photographs in the same
order to the $n$ encoders, allowing the network to optimize each encoder
for each room-type.
In a room-agnostic setting, we randomly change the order of photographs
every time and let all the photograph encoders share the weights.
This matching problem again has ``Yes/No'' answer, and a hinge loss is used.


\subsection{$k$-way matching}
\label{section:K-way-models}
Adding more photographs makes the matching problem easier. Adding more
apartment candidates makes the problem more difficult. The $k$-way
matching problem matches a floorplan to one of $k$ photographs or
$k$ photograph-sets.
%

In a $k$-way photograph matching problem, we concatenate a feature from
a floorplan and $k$ features from photograph encoders that share the
same weights. The concatenated features go through a classification
network consisting of two fully connected layers. The final output is a
one-hot encoding vector of size $k$, indicating which of the $k$
apartments matches the floorplan.
A standard cross entropy loss is used in $k$-way matching problems.
%
The architecture for $k$-way photograph-set matching similarly
concatenates features from all the photographs and all the apartments.
We have not considered room-agnostic matching for $k$-way matching, as
photographs of different room-types exhibit different amount of
information, making the analysis of the matching-accuracy difficult.

\section{Evaluations}


In this section, we first describe the dataset and the implementation
details, then demonstrate how our models perform in a variety of
settings and compete against human vision.

\subsection{Experimental setup}
\label{section:experiments}

\mysubsubsection{Data}
We use the HOME'S dataset~\cite{homes} throughout our experiments. 
It contains data for approximately 5
million apartments in Japan. Each apartment contains one floorplan image
and a set of photographs annotated with room types.
We have selected 100,000 apartments uniformly from the dataset, each of
which has a floorplan image as well as bathroom, kitchen, and
living-room photographs.

For pair matching problems, we form each training pair as a
floorplan and a photograph(-set), either from the same or different apartments.
The ratio of positive to negative samples is 1:1, making random
guess a 50\% chance.
We have generated 99,000 training and 1,000 testing data.

For $k$-way matching problems, each training pair consists of one
floorplan, one matching photograph from the same apartment, and $k-1$
photographs from other apartments randomly sampled from the dataset.
Random guess therefore has a $1/k$ chance.
The numbers of training and testing examples are again 99,000 and 1,000,
respectively.

\mysubsubsection{Implementation details}
We initialize each CNN encoder
as a pretrained VGG16 model. For fine-tuning fully connected layers,
we initialize the weights with a Gaussian function ($\mu = 0$ and
$\sigma = 0.001$). We resize floorplan images to $224 \times 224$
to match the input of the original VGG16 model.
For photographs, their original resolutions are usually
around $100 \times 100$. To save computational expenses, we
resize them to $128 \times 128$ instead of $224 \times
224$. This makes the output of the final max-pooling layer
a $8{,}096 = 4 \times 4 \times 512$ dimensional vector,
instead of $25{,}088 = 7 \times 7 \times 512$. Therefore, we
replace $\it fc6$ with a fully connected layer, which takes
a 8,096 dimensional vector and outputs a 512 dimensional
vector. We change the $\it fc6$ for the floorplan branch to
output also a 512 dimensional vector. All $fc6$ outputs from both floorplan and photographs are concatenated and fed into the following fully connect layer which outputs a 1,024 dimensional vector. The last fully connected layer takes this 1,024 dimensional vector and make the final prediction (similarity score for pair matching and one-hot encoding for $k$-way matching). We
have implemented our model using
Torch7~\cite{collobert2011torch7} and trained our model on
an Nvidia Titan X. Each model takes around 3 days to finish
50 epochs of training.

As discussed in \sect{section:models}, we vary the number of photographs per floorplan and the number of apartments in $k$-way matching, in addition to whether the model is room-aware or room-agnostic. In our experiments, the number of photographs per apartment is either 1 or
3. When set to 1, we choose the room-type of the photograph to be either
a bathroom, a kitchen, or a living-room. When set to 3, these three room
types are used altogether. We set the number of apartments in k-way
matching to either 2, 4, or 8.

\newcommand{\mycyan}[1]{\textcolor{cyan}{#1}}
\newcommand{\myred}[1]{\textcolor{red}{#1}}
\newcommand{\mygreen}[1]{\textcolor{MyDarkGreen}{#1}}
\begin{table*}[t]
  \centering
  \begin{tabular}{l  llll}
   \toprule
   \multirow{2}{*}{\bf Matching type} &
   \multicolumn{4}{c}{\textbf{Photograph type}}\\ 
   \cmidrule{2-5}
                                            & bathroom & kitchen & living room & all\\
    \midrule
   pair-agnostic & $81.2 \pm 2.0$ & $78.8 \pm 3.7$ & $76.0
   \pm 2.8$ & $82.9 \pm 2.1$\\
   pair-aware & $82.3 \pm 1.6$  (\mygreen{51.7}) & $81.8 \pm 2.1$  (\mygreen{58.9})& $77.8
   \pm 1.8$  (\mygreen{59.5}) & $85.3 \pm 3.4$  (\mygreen{61.5})\\
   2-way & $86.2 \pm 1.4$  (\mygreen{64.1}) & $84.8 \pm 3.5$ & $81.2 \pm 1.6$ &$91.0 \pm 1.5$\\
   4-way & $72.4 \pm 3.6$  (\mygreen{43.0}) & $72.4 \pm 1.8$ & $66.5 \pm 1.7$ &$77.8 \pm 2.5$\\
   8-way & $56.9 \pm 1.8$  (\mygreen{42.0}) & $59.3 \pm 1.9$ & $54.0 \pm 3.9$ &$61.4 \pm 2.5$\\
   \bottomrule
  \end{tabular}
  \caption{Matching accuracy. Columns specify the type of
    input photographs. Rows specify the matching problem
    type (pair or k-way and room agnostic or aware). For
    each experiment, we divide the testset into $5$ groups,
    and calculated the average and the standard deviation
    across the 5 groups. The random guess has a chance of
    50\%, 50\%, 25\%, and 12.5\% for the pair, 2-way, 4-way,
    and 8-way problems, respectively. We have also
    conducted the same matching tests with \mygreen{human
      subjects} on Amazon Mechanical Turk, where the green
    numbers show their performance.}
  \label{table:performance}
\end{table*}

\subsection{Results} \label{section:results}

\tbl{table:performance} shows the primary results on our cross-modal
image matching problems. For each of the 20 problem configurations, we
divide the test set into 5 groups, and compute the average accuracy and the standard deviation of accuracy.
Considering the difficulties in our matching problem, it is to our
surprise that the network achieves more than 80\% for most of the pair
matching problems. It is also significantly higher than random guess in
more difficult $k$-way problems.
Comparing the numbers in the top two rows, room-aware networks can
optimize feature encoders for each room type, and outperform
room-agnostic ones consistently.
%
%

\mysubsubsection{Human performance} We have conducted human tests on
Amazon Mechanical Turk for representative problem cases.
For each problem, we have generated 100 questions, and put 10 into
one group. We have repeated the study until we get 3 answers to each of
the question.
%
In order to avoid spammy turkers,
we have copied the first two questions to the end of the group (\ie, 12
questions in total), and only trusted workers who gave
the same answers to these questions.

To our expectation, our matching problem is very challenging and human
performance stays around 50\% for most problems.
Another interesting fact is that it requires 20, 30, and 50 seconds on
average for workers to solve pair/2-way, 4-way, and 8-way
matching problems, respectively. 
This is in contrast to most computer vision problems that
require only instant human reasoning (\ie, one either knows the answer
or not).
In contrast,
our network is able to answer a few dozen questions in a second for all
these cases.
This observation demonstrates that neural networks are also
good at answering questions that require long periods of human
reasoning.

\begin{table}[t]
  \centering
  \begin{tabular}{lcc}
    \toprule
   \multirow{2}{*}{\textbf{Fusion layer}} &
   \multicolumn{2}{c}{\textbf{Fusion function}}   \\ 
   \cmidrule{2-3}
   & Averaging & Concatenation\\
   \midrule
   image & $77.9 \pm 3.2$ & $80.1 \pm 3.2$\\
    $\it conv3$ & $81.0 \pm 2.1$  & $83.1 \pm 2.3$\\
    $\it conv4$ & $82.7 \pm 1.9$  & $83.4 \pm 3.5$\\
    $\it fc6$ & $84.2 \pm 1.8$ & $\mycyan{85.3} \pm 3.4$\\
   score &  $\mycyan{84.7} \pm 2.1$ & $83.3 \pm 2.8$\\
    \bottomrule
  \end{tabular}
  \caption{Performance of different fusion strategies on the set
matching problem. We vary the fusion function and the layer in 
which we fuse the information of photographs. Fusing $\it fc6$ features via concatenation provides the best performance.}  \label{table:set-matching}
\end{table}

\mysubsubsection{Exploiting multiple photographs} As expected, the
network is able to exploit more photographs to improve accuracy (See
\tbl{table:performance}).  We have explored other strategies of
information merging by varying the layer in which we fuse the
information from multiple photographs, either at the layer of images,
convolutional features ($\it conv3$ and $\it conv4$), fully connected
layer features (${\it fc6}$), or predicted scores.
%
%

For the image layer, we fuse 3 photographs into a single 9-channel
photograph. For the layer of convolutional features, we stack all the
feature maps. For the layer of fully connected features, we concatenate
feature vectors.
For the layer of predicted scores, we add one more fully connected layer
that computes their weighted average.

\tbl{table:set-matching} shows matching accuracies over these four
variants as well as the numbers when we simply take the average image,
feature map, feature vector, or score.
As three photographs are not spatially aligned, fusing misaligned
information at an early stage poses unnecessary challenge for the
network. On the other side, fusing only at the end fails to exploit
mutual information properly.
The optimal configuration is to fuse information at the (${\it fc6}$)
layer, where the feature vector encodes the information of an entire
image.

\begin{table}[t]
  \centering
  \begin{tabular}{lc}
    \toprule
   \textbf{Fine-tuning} & \textbf{Accuracy} \\
   \midrule
    none (fixed to VGG) & $69.5 \pm 2.9$\\
    room-agnostic & $83.7 \pm 3.0$\\    
    room-aware (only at fully connected) & $84.1 \pm 1.4$\\
    room-aware & $\mycyan{85.3} \pm 3.4$\\
    \bottomrule
  \end{tabular}
  \caption{Performance on the pair matching problem with the
 photograph-set. We vary whether to supply room type information, and whether and where to finetune the network. Finetuing a room-aware network provides the best performance.}  \label{table:weight-sharing}
\end{table}

\begin{table}[t]
  \centering
  \setlength{\tabcolsep}{5pt}
  \begin{tabular}{lccc}
    \toprule
      \multirow{2}{*}{\bf Training} & \multicolumn{3}{c}{\textbf{Evaluation}} \\ 
   \cmidrule{2-4}
   & 2-way & 4-way & 8-way\\
   \midrule
   pair  & $87.8 \pm 2.6$ & $73.4 \pm 3.6$ & $57.0 \pm 3.3$\\
   2-way & $86.2 \pm 1.4$ &  N/A & N/A \\ 
   4-way & $87.4 \pm 1.7$ & $72.4 \pm 3.6$ & N/A \\ 
   8-way & $86.3 \pm 1.0$ & $71.8 \pm 2.9$ & $56.9 \pm 1.8$\\
    \bottomrule
  \end{tabular}
 \caption{
 Performance of the models trained by different problems.
 The network learns fundamentally the same cross-modal similarity
 metric. We have used the models trained by four different problems
 (rows) to solve the three $k$-way matching problems (columns). We do
 not evaluate models trained on a $k$-way problem for $k^\prime$-way
 problem when $k < k^\prime$,  because it is hard to exclude the
 bias. Please see \sect{section:results} for details.}
 \label{table:K-way-matching}
\end{table}

\begin{figure}[!h]
  \includegraphics[width=\columnwidth]{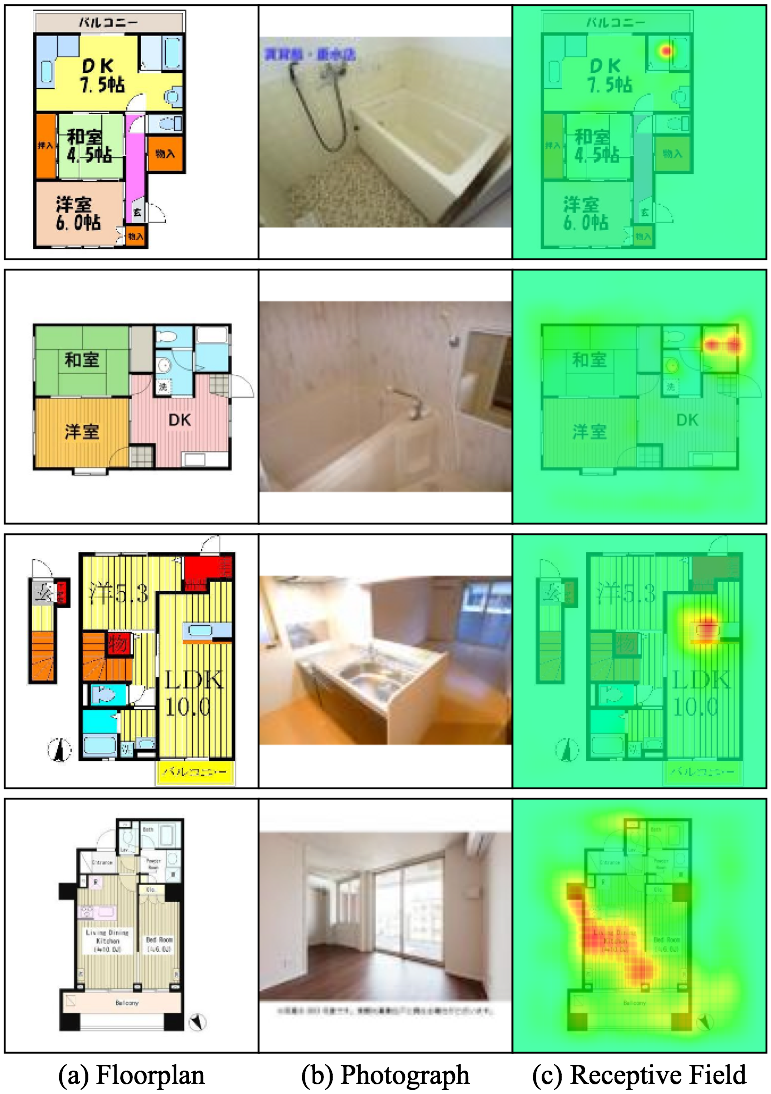}
    \caption{Visualization for the receptive field of the final
      prediction. In each row, from left to right we show an input floorplan, an input photograph, and the corresponding receptive field visualization. Networks learn to localize which part of the floorplan the photograph corresponds to, for bathrooms (the first two rows), kitchens (the third row), and living rooms (the fourth row).}  \label{figure:RF}
\end{figure}

\begin{figure}[!h]
    \includegraphics[width=\columnwidth]{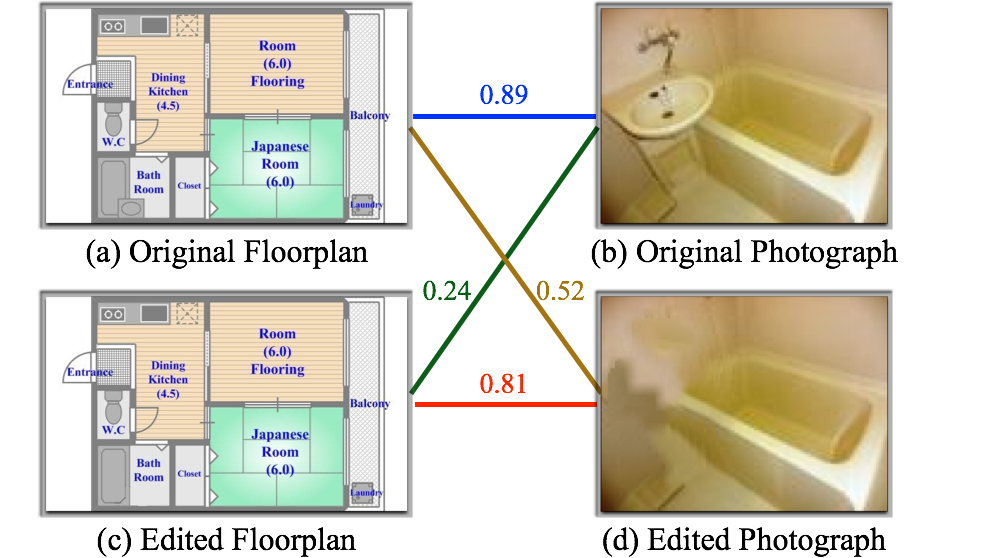}
    \caption{
 The network apparently matches object detections in a floorplan and
 photographs. To verify this hypothesis, we have manually removed the
 washing basin from the floorplan and the photograph (top: original,
 bottom: edited) to look at the four similarity scores highlighted in color.
 }
 \label{figure:object-removal}
\end{figure}

\mysubsubsection{Room-awareness fine-tuning}
The room-aware networks consistently outperform the room-agnostic ones 
as the network can learn separate encoder for each room type.
%
\tbl{table:weight-sharing} studies the effects of room-aware
fine-turning for the pair matching problem with a
photograph-set. The first and the third rows are the new additions to
\tbl{table:performance}.
The first row is a simple baseline where the encoder is
fixed to VGG. The third row shares the encoder weights but
have fully connected layers optimized for each
room type.
The table clearly shows
that the performance improves as the network is given a larger
parameter space for room-specific fine-tuning.



\begin{figure*}[t]
  \includegraphics[width=\textwidth]{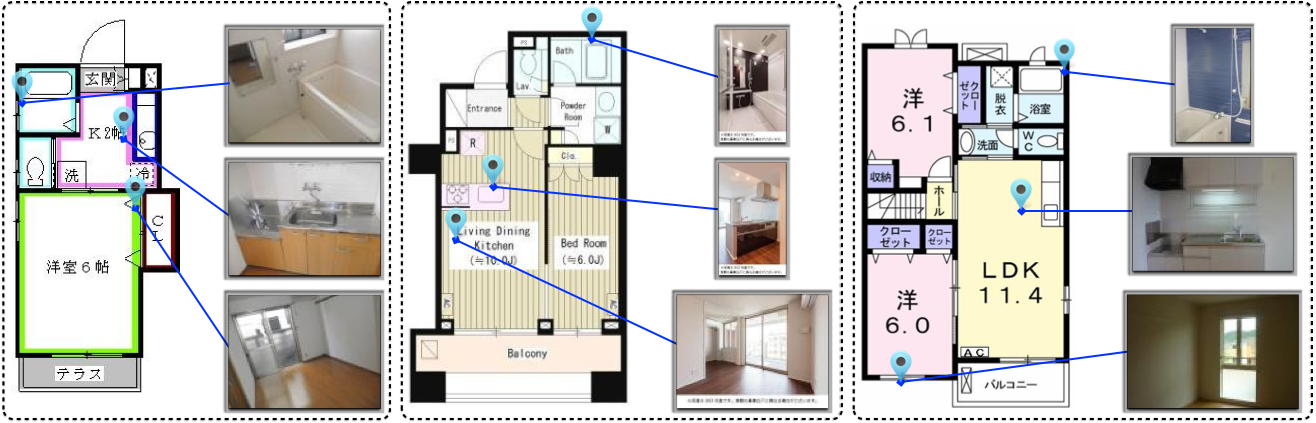}
 \caption{The success of RF visualization allows us to visualize
 photographs in the context of a floorplan image, an effective
 application for real estate websites. We show, for each case, 
 how our model manages to map the kitchen, bathroom, and living room to different
 locations of the floorplan.}
\label{figure:image-placement}
\end{figure*}

\mysubsubsection{Effective learning} The network fundamentally learns
the same similarity metric between a floorplan and a set of photographs
in our family of problems. A natural question is then to ask if one
problem configuration is more effective than others in learning the
metric.  To understand this, we use the $k$-way matching problems to
compare the accuracy of models trained on different problem
configurations
(see \tbl{table:K-way-matching}). More precisely, we have used pair
(room-aware), 2-way, 4-way, and 8-way matching problems with a bathroom
photograph to train networks.

When a model is trained on a pair matching problem, we evaluate the
similarity scores $k$ times to solve the $k$-way matching problems. When
a model is trained on a 8-way problem, we solve 2-way and 4-way problems
by duplicating the input images four times and twice, respectively.
When a model is trained on a 2-way problem, solving 4-way and 8-way
problems without bias is not easy. We thus choose not to include these
cases. Similarly, the model trained on the 4-way problem is used to
solve only the 2-way problem.

The table shows that while trained problems vary in their difficulties,
all the matching accuracies are very similar and do not exhibit
statistically significant differences. This result suggests that, at least
for our problem, one can effectively train a network with the smallest
problem setting, the pair matching problem.

\section{Model interpretation} \label{section:visualization}

\begin{figure*}[t]
  \includegraphics[width=\textwidth]{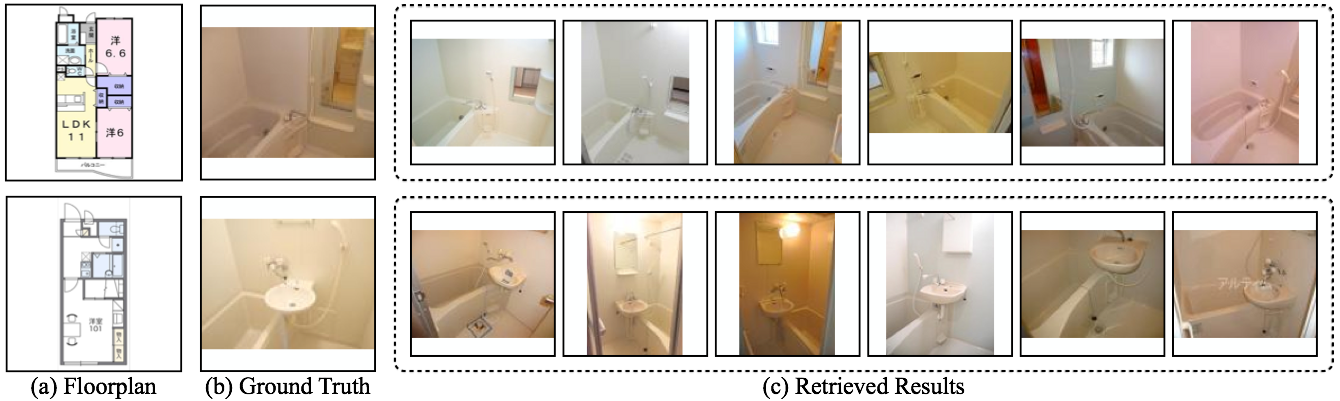}
 \caption{Given a floorplan image, our network can retrieve photographs
 that show likely appearance of the corresponding indoor space. The
 second column shows the ground-truth bathroom photograph, while the
 rest are the top six photographs that have the highest similarity
 scores based on the binary matching network.
}
  \label{figure:image-retrieval}
\end{figure*}

\begin{figure*}[t]
  \includegraphics[width=\linewidth]{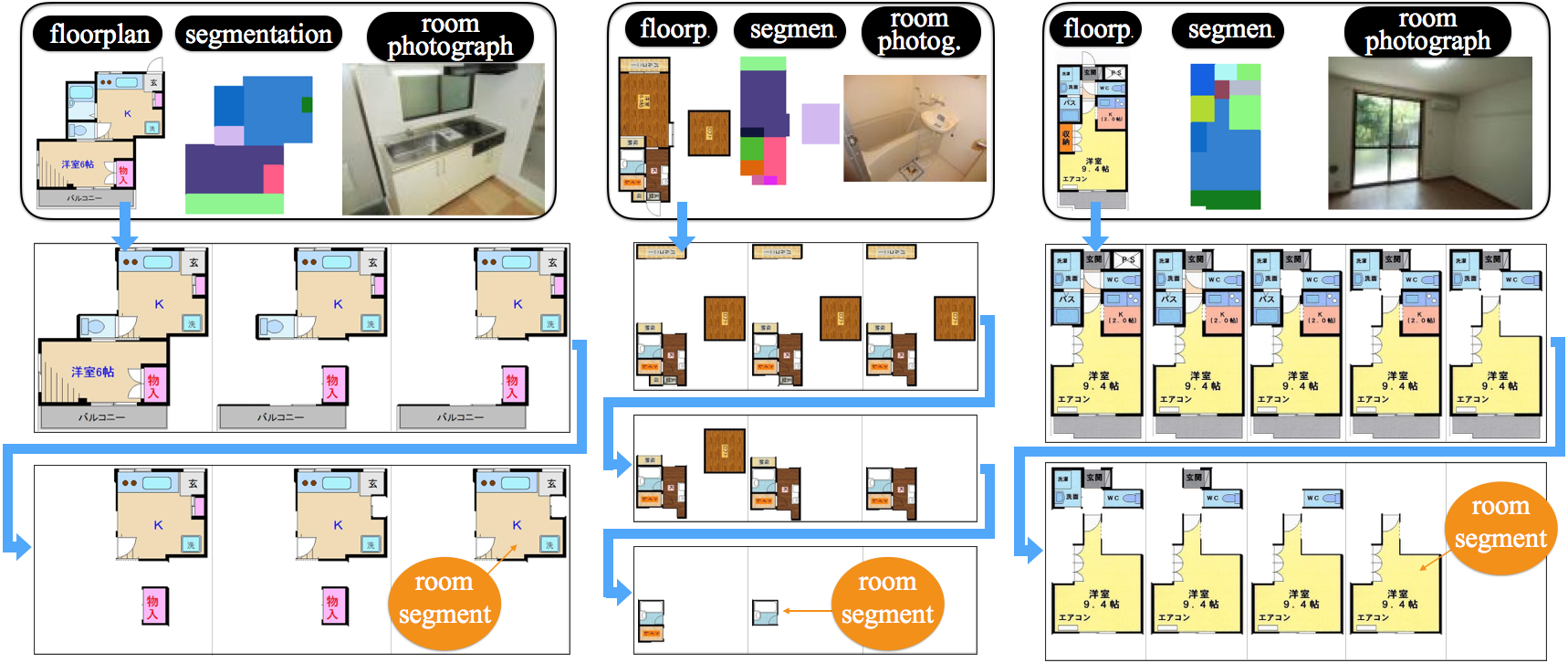}
  \caption{ The simplification
    technique~\cite{zhou2014object} is particularly
    effective for floorplan images, enabling the
    localization of an exact image region corresponding to
    the photograph. Here we show three results side by
    side. The first row contains: left) the floorplan,
    middle) the image, right) the segmentation
    result. Images below show the simplification process
    indicated by blue arrows.
 }
  \label{figure:simplification}
\end{figure*}

Visualization of convolutional networks has been an active area of
research~\cite{zeiler2014visualizing, dosovitskiy2015inverting,
mahendran2015understanding, zhou2014object}. We extend the technique
proposed by Zhou \etal~\cite{zhou2014object} to analyze how our pair
matching network learns to associate floorplan images and photographs
across modalities. We show some results in the main paper, and refer
more to the supplementary material.

\mysubsubsection{Room localization}
We adapt the idea in~\cite{zhou2014object} to visualize the
learned Receptive Fields (RFs).
The difference is that we have two Siamese arms (modalities),
instead of only one for their classification task. Our approach is to
add noise to one arm while fixing the other. More specifically, in a
sliding window manner, we fill a $11\times 11$ window on a floorplan image
with random noise drawn from normal distribution ($\mu=0$ and $\sigma=1$).  The top two examples in \fig{figure:RF} show the
result for a pair matching network with a bathroom photograph. The
third example is with a kitchen photograph and the
fourth is with a living room photograph.
%
The network clearly learns to attend to the corresponding room region in
a floorplan. We have conducted the same visualization test for roughly
50 examples for each room type. We have observed similar
results 40\% of the time.



\mysubsubsection{Object discovery}
The RF visualization indicates that the network learns to attend
to the informative regions on the floorplan to make the
prediction. However, it is still mysterious how the network manages to
achieve such high matching accuracy, much higher than humans. What has
also caught our attention is that the network has consistently recorded
better accuracy with a bathroom or a kitchen over a living room.  Our
hypothesis is that the network learns to detect objects, such as washing
basins, bathtubs, or cooking counters, in both the photograph and the floorplan; 
it then establish correspondences over the detections.
%
%
We validate this hypothesis by manually editing the image and observing how
the similarity score changes. As shown in \fig{figure:object-removal},
at the bottom row,
we have manually edited the floorplan to remove the washing
basin, and have similarly removed it from the photograph by
using PatchMatch software~\cite{barnes2009patchmatch}. The
figure shows the similarity score for every pair, which
clearly indicates that the network uses the presence of an
object and an object-icon to make the prediction.
Please refer to the supplementary material for more examples.

\section{Applications}

In addition to the floorplan-to-photograph matching problem,
the trained networks enable novel applications.

\mysubsubsection{Image placement} Giving a sense of a place to live is a
critical goal of a real estate website. While a floorplan and a set of
photographs serve the purpose to some extent, this is still a
challenging task for real estate websites.
The success of the RF visualization in \sect{section:visualization}
enables a simple but effective algorithm to achieve this aim by placing
photographs or indicating their locations over a floorplan, where the
field response is the maximum (See Fig.\ref{figure:image-placement}).



\mysubsubsection{Image retrieval}
An apartment listing without any photographs is even more difficult to
imagine a sense of a place.
Our network, given a floorplan image, can show likely appearance of the
indoor space through image retrieval. More precisely, we use a pair
matching network to identify photographs with high similarity
scores. \fig{figure:image-retrieval} shows the top six bathroom
photographs with the highest similarity scores with ground truth on the
left. Notice that all retrieved photographs exhibit consistent
appearance and content.
%

\mysubsubsection{Localization}
The simplification visualization technique in~\cite{zhou2014object}
suits our problem perfectly, since image segmentation is highly effective for
floorplans that originate from vector graphics.
The technique allows us to localize an exact image region that
corresponds to a photograph in addition to its rough location.
We use Photoshop to segment the floorplan image as shown in
\fig{figure:simplification}.
We then repeat removing a segment that has the least change in the
similarity score.
As \fig{figure:simplification} demonstrates, this simplification process
often produces correct image segments corresponding to the room of a photograph.

\section{Conclusions}

This paper has introduced a novel multi-modal image correspondence
problem with human baselines.  This is a very challenging problem that
requires long periods of reasoning even for humans, unlike other
conventional computer vision problems that only require instant human
attention.  We have explored various deep network architectures for this
task, and demonstrated that they achieve high matching accuracies,
significantly outperforming human subjects. We have conducted a wide
range of qualitative and quantitative experiments, and analyzed and
visualized the learned representation.  Lastly, we have shown a few
applications utilizing the power of trained networks which have been
otherwise impossible. We believe that this paper provides a new insight
in the machine vision's capability of cross-modal image matching, and
promotes future research in this under-explored domain.




\section{Acknowledgement} \label{section:acknowledgement}
This research is partially supported by National Science
Foundation under grant IIS 1540012 and IIS 1618685, Google
Faculty Research Award, and Microsoft Azure Research
Award. Jiajun Wu is supported by an Nvidia fellowship. This
research was partially conducted while Jiajun Wu was
interning at Microsoft Research. We thank Nvidia for a
generous GPU donation.

\clearpage

\clearpage
{
  \small
  \bibliographystyle{ieee}
  \bibliography{ref}

\begin{thebibliography}{10}\itemsep=-1pt

\bibitem{homes}
{HOME'S} dataset.
\newblock \url{http://www.nii.ac.jp/dsc/idr/next/homes.html}.

\bibitem{altwaijry2016learning}
H.~Altwaijry, E.~Trulls, J.~Hays, P.~Fua, and S.~Belongie.
\newblock Learning to match aerial images with deep attentive architectures.
\newblock In {\em CVPR}, 2016.

\bibitem{antol2015vqa}
S.~Antol, A.~Agrawal, J.~Lu, M.~Mitchell, D.~Batra, C.~Lawrence~Zitnick, and
  D.~Parikh.
\newblock Vqa: Visual question answering.
\newblock In {\em ICCV}, 2015.

\bibitem{aytar2016soundnet}
Y.~Aytar, C.~Vondrick, and A.~Torralba.
\newblock Soundnet: Learning sound representations from unlabeled video.
\newblock In {\em NIPS}, 2016.

\bibitem{bansal2012ultra}
M.~Bansal, K.~Daniilidis, and H.~Sawhney.
\newblock Ultra-wide baseline facade matching for geo-localization.
\newblock In {\em ECCV}, 2012.

\bibitem{barnes2009patchmatch}
C.~Barnes, E.~Shechtman, A.~Finkelstein, and D.~Goldman.
\newblock Patchmatch: a randomized correspondence algorithm for structural
  image editing.
\newblock {\em ACM TOG}, 28(3):24, 2009.

\bibitem{bay2006surf}
H.~Bay, T.~Tuytelaars, and L.~Van~Gool.
\newblock Surf: Speeded up robust features.
\newblock In {\em ECCV}, 2006.

\bibitem{bernardi2016automatic}
R.~Bernardi, R.~Cakici, D.~Elliott, A.~Erdem, E.~Erdem, N.~Ikizler-Cinbis,
  F.~Keller, A.~Muscat, and B.~Plank.
\newblock Automatic description generation from images: A survey of models,
  datasets, and evaluation measures.
\newblock {\em JAIR}, 55:409--442, 2016.

\bibitem{bromley1993signature}
J.~Bromley, J.~W. Bentz, L.~Bottou, I.~Guyon, Y.~LeCun, C.~Moore,
  E.~S{\"a}ckinger, and R.~Shah.
\newblock Signature verification using a “siamese” time delay neural
  network.
\newblock {\em IJPRAI}, 7(04):669--688, 1993.

\bibitem{castrejonlearning}
L.~Castrej{\'o}n, Y.~Aytar, C.~Vondrick, H.~Pirsiavash, and A.~Torralba.
\newblock Learning aligned cross-modal representations from weakly aligned
  data.
\newblock In {\em CVPR}, 2006.

\bibitem{chu2015you}
H.~Chu, D.~Ki~Kim, and T.~Chen.
\newblock You are here: Mimicking the human thinking process in reading
  floor-plans.
\newblock In {\em ICCV}, 2015.

\bibitem{collobert2011torch7}
R.~Collobert, K.~Kavukcuoglu, and C.~Farabet.
\newblock Torch7: A matlab-like environment for machine learning.
\newblock In {\em BigLearn, NIPS Workshop}, 2011.

\bibitem{dosovitskiy2015inverting}
A.~Dosovitskiy and T.~Brox.
\newblock Inverting convolutional networks with convolutional networks.
\newblock In {\em CVPR}, 2016.

\bibitem{han2015matchnet}
X.~Han, T.~Leung, Y.~Jia, R.~Sukthankar, and A.~C. Berg.
\newblock Matchnet: Unifying feature and metric learning for patch-based
  matching.
\newblock In {\em CVPR}, 2015.

\bibitem{huang2016visual}
T.-H.~K. Huang, F.~Ferraro, N.~Mostafazadeh, I.~Misra, A.~Agrawal, J.~Devlin,
  R.~Girshick, X.~He, P.~Kohli, D.~Batra, et~al.
\newblock Visual storytelling.
\newblock In {\em NAACL}, 2016.

\bibitem{jaderberg2015spatial}
M.~Jaderberg, K.~Simonyan, A.~Zisserman, et~al.
\newblock Spatial transformer networks.
\newblock In {\em NIPS}, 2015.

\bibitem{karpathy2015deep}
A.~Karpathy and L.~Fei-Fei.
\newblock Deep visual-semantic alignments for generating image descriptions.
\newblock In {\em CVPR}, 2015.

\bibitem{lin2013cross}
T.-Y. Lin, S.~Belongie, and J.~Hays.
\newblock Cross-view image geolocalization.
\newblock In {\em CVPR}, 2013.

\bibitem{lin2015learning}
T.-Y. Lin, Y.~Cui, S.~Belongie, and J.~Hays.
\newblock Learning deep representations for ground-to-aerial geolocalization.
\newblock In {\em CVPR}, 2015.

\bibitem{liu2015rent3d}
C.~Liu, A.~G. Schwing, K.~Kundu, R.~Urtasun, and S.~Fidler.
\newblock Rent3d: Floor-plan priors for monocular layout estimation.
\newblock In {\em CVPR}, 2015.

\bibitem{long2014convnets}
J.~L. Long, N.~Zhang, and T.~Darrell.
\newblock Do convnets learn correspondence?
\newblock In {\em NIPS}, 2014.

\bibitem{lowe1999object}
D.~G. Lowe.
\newblock Object recognition from local scale-invariant features.
\newblock In {\em ICCV}, 1999.

\bibitem{mahendran2015understanding}
A.~Mahendran and A.~Vedaldi.
\newblock Understanding deep image representations by inverting them.
\newblock In {\em CVPR}, 2015.

\bibitem{martin20143d}
R.~Martin-Brualla, Y.~He, B.~C. Russell, and S.~M. Seitz.
\newblock The 3d jigsaw puzzle: Mapping large indoor spaces.
\newblock In {\em ECCV}, 2014.

\bibitem{owens2016ambient}
A.~Owens, J.~Wu, J.~H. McDermott, W.~T. Freeman, and A.~Torralba.
\newblock Ambient sound provides supervision for visual learning.
\newblock In {\em ECCV}, 2016.

\bibitem{simo2015discriminative}
E.~Simo-Serra, E.~Trulls, L.~Ferraz, I.~Kokkinos, P.~Fua, and F.~Moreno-Noguer.
\newblock Discriminative learning of deep convolutional feature point
  descriptors.
\newblock In {\em ICCV}, 2015.

\bibitem{simonyan2014very}
K.~Simonyan and A.~Zisserman.
\newblock Very deep convolutional networks for large-scale image recognition.
\newblock {\em arXiv preprint arXiv:1409.1556}, 2014.

\bibitem{wang2015lost}
S.~Wang, S.~Fidler, and R.~Urtasun.
\newblock Lost shopping! monocular localization in large indoor spaces.
\newblock In {\em ICCV}, 2015.

\bibitem{wolff2016regularity}
M.~Wolff, R.~T. Collins, and Y.~Liu.
\newblock Regularity-driven facade matching between aerial and street views.
\newblock In {\em CVPR}, 2016.

\bibitem{zagoruyko2015learning}
S.~Zagoruyko and N.~Komodakis.
\newblock Learning to compare image patches via convolutional neural networks.
\newblock In {\em CVPR}, 2015.

\bibitem{zeiler2014visualizing}
M.~D. Zeiler and R.~Fergus.
\newblock Visualizing and understanding convolutional networks.
\newblock In {\em ECCV}, 2014.

\bibitem{zhou2014object}
B.~Zhou, A.~Khosla, A.~Lapedriza, A.~Oliva, and A.~Torralba.
\newblock Object detectors emerge in deep scene cnns.
\newblock In {\em ICLR}, 2015.

\end{thebibliography}
}

\end{document}